\documentclass{article}

\usepackage{amsmath}        
\usepackage{amssymb}        
\usepackage{bm}             
\usepackage{siunitx}        

\usepackage{graphicx}       
\usepackage{booktabs}       
\usepackage{multirow}       

\usepackage{hyperref}       
\hypersetup{
    colorlinks=true,
    linkcolor=blue,
    filecolor=magenta,      
    urlcolor=cyan,
    citecolor=blue,         
}
\usepackage{cleveref}       

\usepackage{xcolor}  

\usepackage{graphicx}
\usepackage{multirow}
\usepackage{amsmath,amssymb,amsfonts}
\usepackage{amsthm}
\usepackage{mathrsfs}
\usepackage[title]{appendix}
\usepackage{xcolor}
\usepackage{textcomp}
\usepackage{manyfoot}
\usepackage{booktabs}
\usepackage{algorithm}
\usepackage{algorithmicx}
\usepackage{algpseudocode}
\usepackage{listings}
\usepackage[margin=1in]{geometry}
\usepackage{authblk} 
\usepackage[natbib=true,style=numeric,sorting=none]{biblatex}
\usepackage[font=small]{caption} 
\usepackage{xspace}
\usepackage{fltpage}

\makeatletter
\def\ps@plain{%
  \let\@oddhead\@empty
  \let\@evenhead\@empty
  \def\@oddfoot{\hfill\raisebox{-0.5cm}{\makebox[0pt][r]{\thepage}}}%
  \def\@evenfoot{\@oddfoot}%
}
\makeatother
\def\ourmethod{MicroDiffuse3D\xspace}

{}

\begin{document}

\title{\ourmethod: A Foundation Model for 3D Microscopy Imaging Restoration}
\author[1]{Yongkang Li}
\author[2]{Brian Wong}
\author[2]{King Wai Chiu}
\author[1]{Hanwen Xu}
\author[1]{Tangqi Fang}
\author[2]{Erin Dunnington}
\author[2]{Dan Fu$^*$}
\author[1]{Sheng Wang\footnote{Correspondence:  danfu@uw.edu, swang@cs.washington.edu}}
\affil[1]{Paul G. Allen School of Computer Science and Engineering, University of Washington, Seattle, WA, USA.}
\affil[2]{Department of Chemistry, University of Washington, Seattle, WA, USA}
\date{}

\maketitle

\begin{abstract}

Chemical imaging enables label-free visualization of cells, tissues and living systems while providing direct biochemical information that is difficult to obtain with conventional fluorescence microscopy. Despite its promise in applications ranging from intraoperative diagnosis to drug-response analysis, its broader use remains limited by slow data acquisition, particularly for three-dimensional imaging. In practice, imaging large volumes or many samples at high spatial resolution is often prohibitively slow, creating a major throughput bottleneck for biomedical studies. Computational super-resolution is a promising approach to break the tradeoff between resolution and image volume. Although many methods have been developed for microscopy image restoration and enhancement, the problem of recovering high-resolution 3D structure from throughput-optimized low-resolution chemical imaging measurements has remained largely unexplored.
Here we present \ourmethod, a pretrained foundation model for 3D microscopy image restoration that recovers high-quality volumetric structure from degraded low-resolution measurements acquired at substantially higher throughput. Built on large-scale pretraining over a curated corpus of 2.55 million microscopy images, \ourmethod combines broad biological and spatial structural priors learned from the data with strong generative capabilities of diffusion-based reconstruction. By restoring volumes jointly rather than slice by slice, the model better recovers sharp cellular structures in 3D while preserving consistency with the measured signal.
We evaluated \ourmethod across three challenging restoration settings, including 3D super-resolution under 16-fold volumetric sparsity, joint degradation in resolution and noise, and 3D denoising in the low signal-to-noise ratio (SNR) regime. The model delivered clear gains over strong baselines in the two super-resolution-related settings, while remaining competitive in 3D denoising against methods specifically engineered for that task. Under the sparse 3D super-resolution setting, \ourmethod produced clearer continuity across depth with fewer artifacts and improved segmentation quality by 10.58\% and line-profile concordance by 15.59\%.
Together, our results establish pretrained 3D restoration as a broadly applicable strategy for overcoming the throughput and SNR limitations in volumetric chemical imaging, enabling high-resolution analysis at scales and speeds that were previously difficult to achieve.

\end{abstract}

\newpage

\section*{Introduction}

Chemical imaging has emerged as a powerful tool in biomedicine because it enables label-free visualization of cells, tissues, and even live animals at subcellular resolution. Compared to widely used fluorescence microscopy, popular chemical imaging modalities such as stimulated Raman scattering (SRS) microscopy~\cite{Freudiger_Science, Cheng_Science} and optical photothermal infrared (O-PTIR) imaging~\cite{Bai_ScienceAdvances, Cheng_NatMet} provide noninvasive analysis of biochemical composition and tracking of both native biomolecules and exogenous molecules such as metabolic tracers and small molecule drugs. They have shown tremendous value in applications ranging from guiding intraoperative diagnosis~\cite{Orringer_NatBiomedEng,Hollon_NatMed}, probing metabolic heterogeneity~\cite{Cheng_SciAdv1,Shi_NatComm,Min_NatBioMed}, to tracing single cell drug transport and response~\cite{Wong_AnalChem,Xu_AnalChem}, all without exogenous labeling. Yet most of these applications are still constrained by the low throughput of chemical imaging, limiting the size or number of samples that can be analyzed within reasonable timeframe. For example, regular SRS imaging can achieve $\sim300\space nm$ lateral resolution and $1\space \mu m$ axial resolution, but typical field of view is only $\sim200\space\mu m$, resulting in a space–bandwidth product that is nearly two orders of magnitude lower than widefield fluorescence microscopy platforms~\cite{Manifold_Annal,Gao_AdvPhotonics}. As a result, large area volumetric acquisition becomes prohibitively slow.

This bottleneck arises from the acquisition physics of chemical imaging. Most chemical imaging systems are point-scanning and sensitivity-limited, resulting in a fundamental tradeoff between spatial resolution, field of view, and acquisition speed. Pixel dwell times typically ranges from 2-100 us, constrained by the shot noise of the detected laser beam~\cite{Min_SciAdv}. As a result, high-resolution imaging over large fields or 3D volumes is often prohibitively slow. Computational super-resolution offers an attractive route to alleviate this tradeoff by acquiring lower-resolution data over a larger field quickly and reconstructing higher-resolution outputs. Prior computational microscopy work have addressed related restoration problems from several angles, including denoising, deconvolution, volumetric super-resolution, and axial resolution enhancement\cite{care_weigert,crossmodality_wang,dfcan_qiao,zsdeconvnet_qiao,rln_li, 3drcan_chen, axialinr_kyungryun, park_park, selfnet_ning}. More recently, approaches such as InterpolAI improve volumetric throughput by inferring missing slices from neighboring observations~\cite{interpolai_joshi}. Unfortunately, these approaches have not yet been adopted in volumetric chemical imaging, in part due to the lack of large-scale datasets and differences in image quality and contrast. The central practical goal in chemical imaging is throughput improvement: computationally reconstructing high-resolution 3D structures from fast, lower-resolution measurements covering a larger field of view. These measurements suffer from not only anisotropic lateral and axial blurring (axial resolution degrades faster than lateral resolution), but also degradation in signal to noise ratio (SNR) (\textbf{Supplementary Fig.}~\ref{suppfig:1}). Addressing this problem could transform biomedical applications such as surgical margin detection or large-scale drug screening, where throughput is a key constraint. This specific regime remains unexplored in chemical imaging, in contrast to the established focus on image restoration, missing slice prediction, or anisotropy correction in fluorescence microscopy. Given the diverse contrast mechanisms and the cost of curating paired volumetric datasets, a robust and general-purpose 3D foundation model - rather than task-specific designs - is needed to reconstruct complex volumetric degradations. 


Here, we propose \ourmethod, a 3D foundation model for recovering high-resolution volumetric structure from severely degraded chemical imaging data. The core idea is to couple large-scale pretraining with conditional 3D diffusion reconstruction. The model learns a strong prior over volumetric structure from pretraining on a large scale 3D SRS imaging data, then leverages that prior at inference through a conditional 3D diffusion transformer to reconstruct target resolution volumes that remain faithful to the acquired signal.  The model combines decomposed attention with condition injection, allowing it to enforce measurement constraints while aggregating long-range spatial context throughout the volume. Unlike methods that process slices independently or interpolating missing content between sampled planes, \ourmethod directly restores the full 3D volume at target-resolution, which recovers axial continuity and fine structural detail from severely degraded measurements.


We evaluated \ourmethod across three challenging settings: 3D super-resolution of SRS microscopy images under 16-fold volumetric sparsity, BioTISR (Biological Time-lapse Image Super-Resolution) image restoration with joint degradation in resolution and noise, and 3D Denoise of SRS microscopy images at SNR $< 5$.
First, we demonstrated that \ourmethod achieves consistent gains over established baselines, with average improvements of +0.68 dB in Peak Signal-to-Noise Ratio (PSNR), +3.73\% in Structural Similarity (SSIM~\cite{ssim_zhou}), +8.22\% in Multi-Scale Structural Similarity (MS-SSIM~\cite{msssim_zhou}), and -7.92\% in Learned Perceptual Image Patch Similarity (LPIPS~\cite{lpips_richard}) over the most competitive baseline across the two SR-related datasets. 
Notably, we observed substantially improved axial fidelity with clearer cross-depth continuity and fewer overlap artifacts in orthogonal views. 
Second, in the 3D Denoising task, \ourmethod delivered restoration quality comparable to state-of-the-art methods specifically engineered for volumetric denoising, despite being a more broadly applicable framework.
Third, under the 3D SR(SRS) 16-fold sparsity setting, we observed that the volumes restored by our model lead to significant gains in segmentation quality (Panoptic Quality, +10.58\%) and line-profile concordance (Lin’s concordance correlation coefficient, +15.59\%), showing that improved recovery quality translates into more reliable downstream biological quantification. 

In summary, \ourmethod is a pretrained 3D diffusion transformer that bridges the gap between high-throughput acquisition and high-resolution demands of downstream analysis while establishing a versatile framework for general volumetric image restoration. By combining large-scale pretraining with conditional 3D diffusion reconstruction, \ourmethod achieves faithful volumetric recovery across challenging restoration settings and supports reliable downstream biological quantification.
\newpage
\section*{Results}\label{sec:results}
\subsection*{\ourmethod: A Foundation Model for Volumetric Restoration}
\begin{figure}
    \centering
    \includegraphics[width=\textwidth]{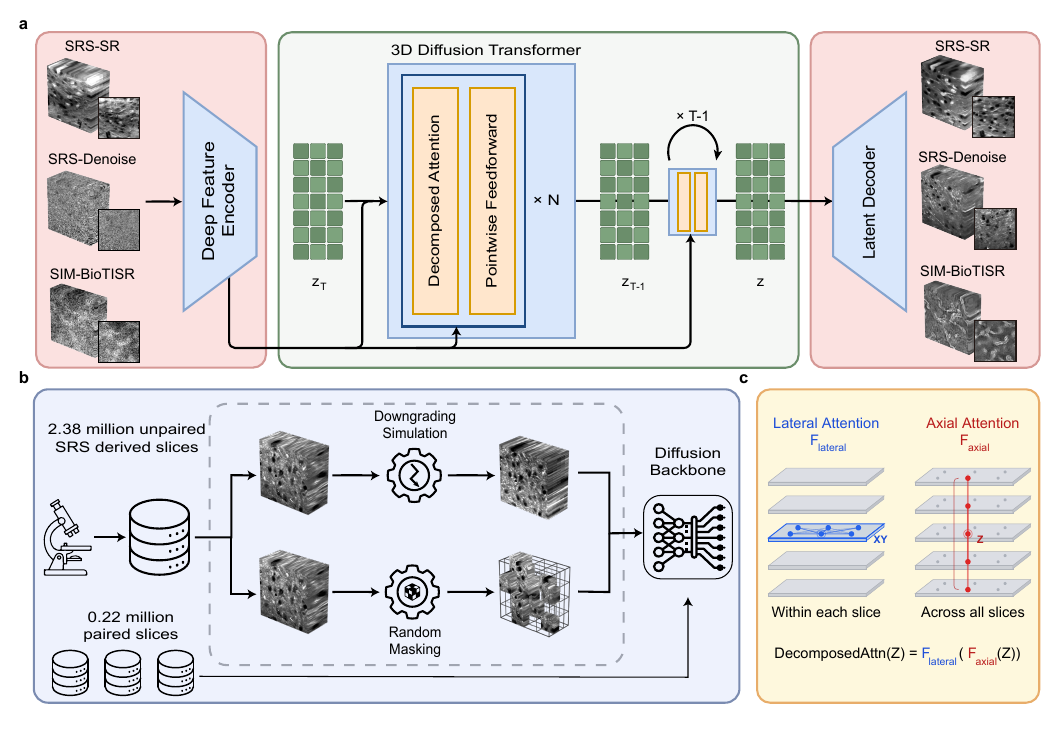}
    \caption{\textbf{Overview of \ourmethod.} \textbf{a, Conditional Diffusion Architecture.} \ourmethod takes anisotropic volumetric data as condition and processes them via a deep feature encoder. Conditioning features are then injected into the latent diffusion process for high-fidelity reconstruction. \textbf{b, Large-scale pretraining strategy.} The model is pretrained on 2.38 million unpaired SRS-derived slices using two complementary objectives: physics-informed downgrading simulation and random 3D masking. These pretraining tasks expose the diffusion backbone to realistic degradations and missing-information recovery before downstream task-specific finetuning. \textbf{c, Decomposed attention design.} Decomposed attention is composed of lateral attention and axial attention. Lateral attention models within-slice spatial structure, whereas axial attention captures dependencies across slices, enabling efficient volumetric context modeling.}
    \label{fig1}
\end{figure}

To construct a generative foundation for computational microscopy, \ourmethod leverages a massive, dimensionally heterogeneous dataset comprising 2.38 million unpaired spatial data acquired via high-resolution SRS microscopy (in number of slices, standardized to $256 \times 256$ spatial dimensions). In addition, paired volumes (0.17 million, in number of high-quality groundtruth slices) acquired from SRS microscopy and the public BioTISR dataset are also fed into the pretraining stage. Our model encompasses a two-stage paradigm: a self-supervised pretraining phase (\textbf{Fig.} \ref{fig1}\textbf{b}) followed by task-specific finetuning. During pretraining, this expansive unpaired corpus is subjected to dynamic physical degradation simulations and 3D block-masking, explicitly forcing the network to internalize a biological manifold and robust structural priors. 
For downstream applications, the model undergoes supervised finetuning utilizing paired volumes of low-quality inputs and high-resolution targets for supervised adaptation. Ultimately, in the inference stage, the model relies solely on degraded volumes as conditional inputs to guide high-resolution volumetric restoration.

\ourmethod operates as a transformer-based Latent Diffusion Model that achieves high-fidelity volumetric reconstruction through a rigorous global-local dual-conditioning mechanism (\textbf{Fig} \ref{fig1}\textbf{a}). This formulation drives a native 3D generative process, utilizing attention within the latent space to synthesize continuous, anatomically realistic biological structures. To address the unique geometric challenges of volumetric data, we build on the Diffusion Transformer (DiT)~\cite{dit_william,sit_peebles} and specifically adapt it to a 3D backbone. This improved architecture naturally processes 3D patch sequences. By utilizing decomposed lateral (XY-plane) and axial (Z-axis) attention mechanisms, this architecture seamlessly accommodates the optical anisotropy of microscopic imaging while circumventing the prohibitive memory costs of full 3D attention.

We applied this pretrained foundation model across multiple downstream restoration tasks, with a primary focus on 3D super-resolution under extreme volumetric sparsity (e.g. a 16-fold undersampling comprising 4-fold in axial and 4-fold in lateral degradation). 
We then tested the model in a Structured Illumination Microscopy (SIM) setting using the public BioTISR (Biological Time-lapse Image Super-Resolution)~\cite{BIOTISR_qiao} dataset, where restoration must be performed under joint degradation in both noise and resolution.
In addition, we evaluated performance in a second restoration regime, extreme-noise 3D denoising at signal-to-noise ratios below 5. 
Furthermore, we examined whether improved volumetric super-resolution translated into more reliable downstream biological quantification, including nuclei segmentation and line-profile concordance.

\subsection*{3D Super-resolution for SRS and BioTISR}

\begin{figure}
    \centering
    \includegraphics[width=\textwidth]{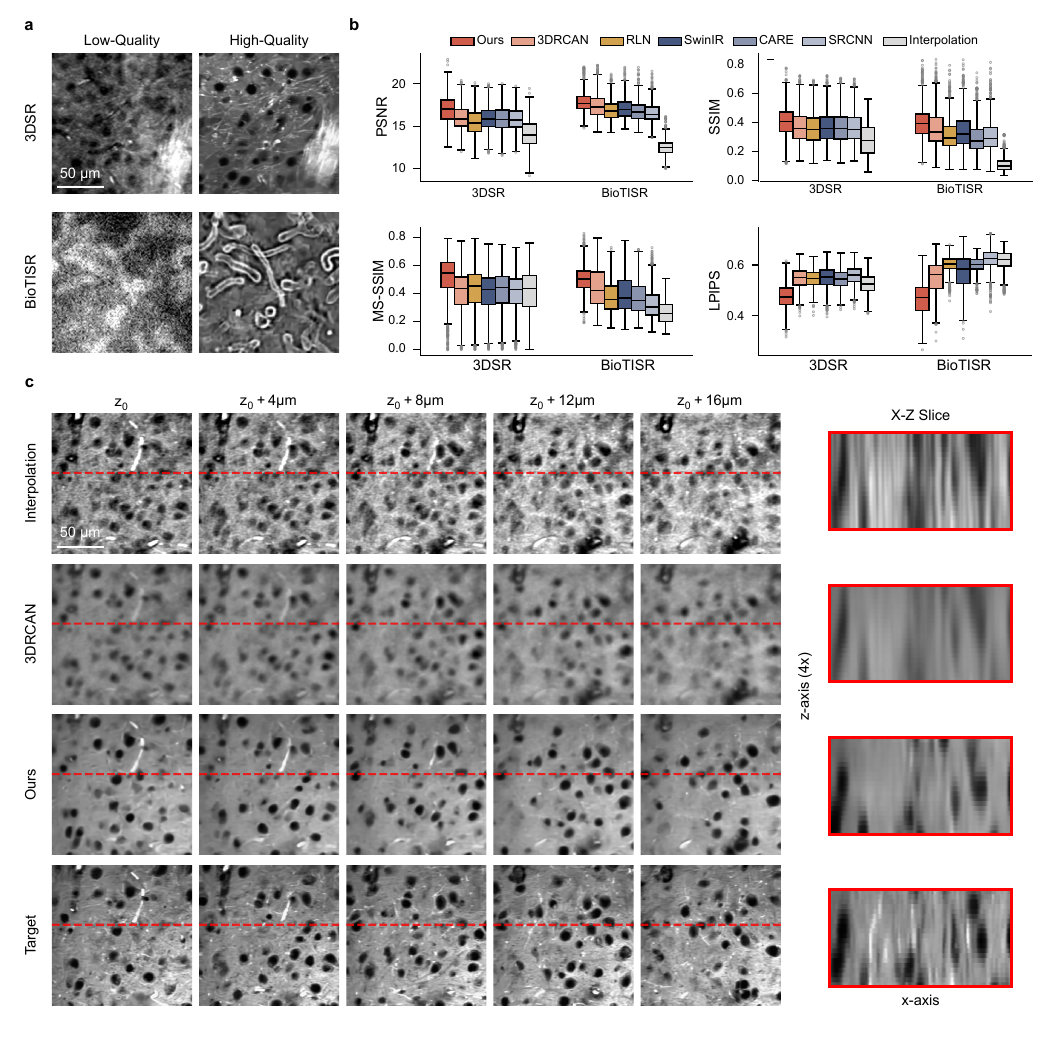}
    \caption{\textbf{3D super-resolution across microscopy modalities.} \textbf{a, b. Quantitative evaluation of super-resolution performance.} Box plots detail PSNR, SSIM, MS-SSIM and LPIPS distributions across independent volumetric samples from SIM and SRS datasets. Our 3D diffusion model consistently outperforms baselines. \textbf{c. Qualitative assessment of axial structural fidelity.} Representative volumetric cross-sections demonstrate that existing models (represented by 3DRCAN, Row 2) fail to disentangle dense 3D spatial information, suffering from overlap artifacts along the Z-axis. In contrast, \ourmethod (Row 3) successfully resolves axial biological structures.
    }
    \label{fig2}
\end{figure}


We first evaluated MicroDiffuse3D on volumetric super-resolution tasks where low resolution images are restored into high resolution images, effectively boosting the volumetric throughput of the microscope. To test the universality of the framework, evaluations were conducted across two fundamentally distinct microscopy modalities: Structured Illumination Microscopy (SIM) and SRS microscopy. 

\ourmethod consistently outperformed the baselines across all evaluated metrics (\textbf{Fig.}\ref{fig2}\textbf{b}). Our model achieves significant margins in both structural similarity (SSIM / MS-SSIM) and peak signal-to-noise ratio (PSNR) compared to established baselines, confirming highly accurate pixel-level and structural-level reconstruction. Furthermore, the marked gains in Learned Perceptual Image Patch Similarity (LPIPS) indicate that our reconstructions are perceptually aligned with ground-truth biological textures. The consistency of these performance gains in both SIM and SRS datasets with vastly different features supports the value of diffusion-based 3D modeling for volumetric reconstruction across distinct imaging regimes.

Beyond numerical metrics, the axial views (\textbf{Fig.}~\ref{fig2}\textbf{c}) and enlarged lateral views (\textbf{Supplementary Fig.}~\ref{suppfig:2}) reveal a consistent qualitative difference. Baselines fail to disentangle dense 3D spatial information. They often produce cross-plane overlap artifacts and blurred axial boundaries. In contrast, \ourmethod recovers more coherent and faithful structures across depth, consistent with the intended role of cross-slice conditioning and decomposed spatial–depth attention. 

Furthermore, exact registration in the 3D SR setting is fundamentally constrained by physical acquisition limits, as minor lateral and axial shifts introduce spatial deviations into the paired datasets. Compounded with the 16-fold volumetric sparsity, recovering pixel-perfect morphology is a highly demanding task. It is unavoidable to have minor spatial deviations and some loss of high-frequency details.

\subsection*{3D Image Restoration under Extreme Noise}

\begin{figure}
    \centering
    \includegraphics[width=\textwidth]{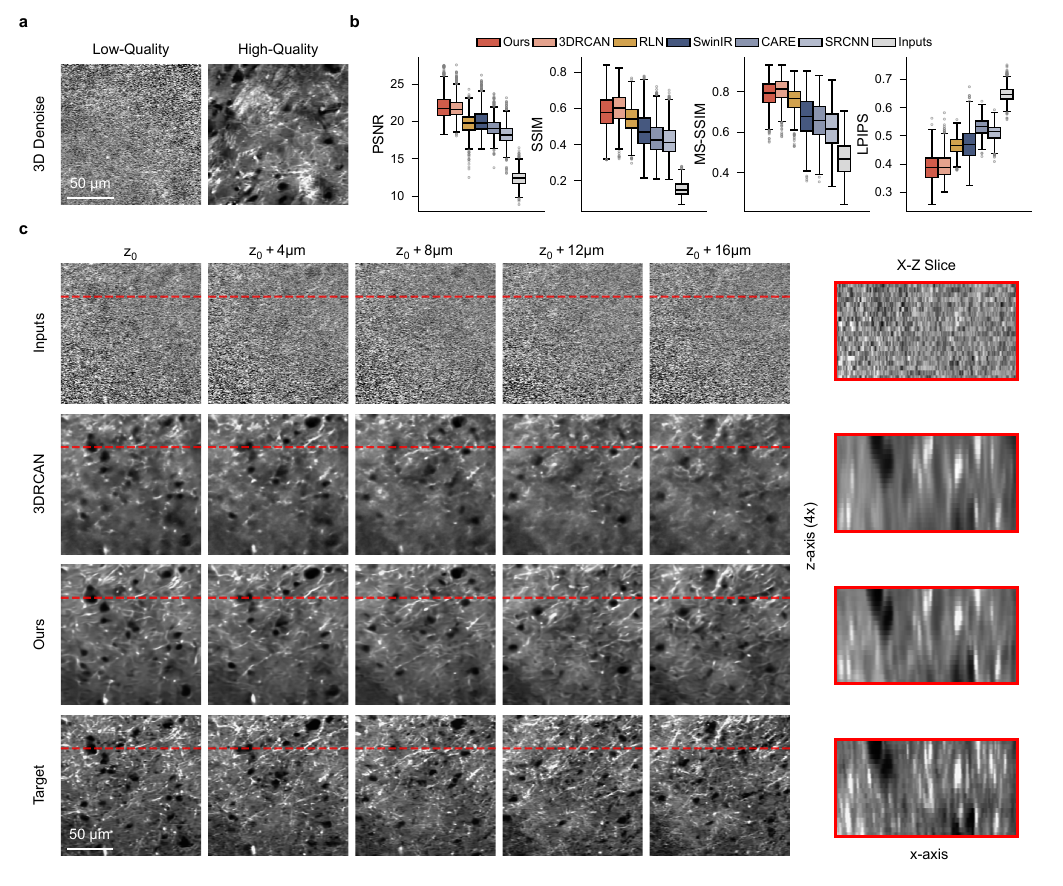}
    \caption{\textbf{3D Denoising.} \textbf{a, b. Quantitative evaluation of denoising performance. }Box plots detail PSNR, SSIM, MS-SSIM and LPIPS distributions across independent volumetric samples from SRS dataset. Our 3D diffusion model consistently outperforms 2D-based baselines.
    \textbf{c. Qualitative assessment of recovery and axial structural fidelity.} Representative cross-sections of baselines (represented by 3DRCAN, Row 2) and our model (Row 3) demonstrate capability of signal recovery and cross-depth continuity preservation. 
    }
    \label{fig3}
\end{figure}

We next evaluated whether the same framework remains effective under a distinct but related restoration challenge: 3D denoising at extremely low signal-to-noise ratios. As imaging depth into biological tissue increases, severe optical scattering and limited photon budgets inevitably degrade image quality, often reducing the SNR to extreme levels (e.g., SNR $\leq$ 5). In this regime, the effective biological signal within any single slice is severely limited, therefore successful recovery depends less on within-slice enhancement and more on aggregating corroborating features across neighboring planes. 

Quantitative evaluation on the SRS denoising dataset (\textbf{Fig.}~\ref{fig3}\textbf{b}) shows that \ourmethod\ remains highly competitive in this demanding setting. Although it is not specifically engineered for volumetric denoising, \ourmethod achieves restoration quality comparable to state-of-the-art methods designed for this task, while significantly outperforming conventional slice-wise restoration baselines across multiple metrics. In particular, its strong performance in PSNR and SSIM indicates effective recovery of underlying biological signal under severe noise corruption, whereas its favorable MS-SSIM and LPIPS distributions suggest preservation of multi-scale structural detail. 

The qualitative assessment (\textbf{Fig.}~\ref{fig3}\textbf{c}, \textbf{Supplementary Fig.}~\ref{suppfig:3}) further supports this interpretation. Representative lateral views and cross-sections show that  \ourmethod achieve clean signal recovery and maintain coherent structural continuity in the restored volumes. Notably, comparison with the baseline methods, \ourmethod produces sharper anatomical features (evidenced by higher energy in the high-frequency spectrum and higher Laplacian variance, \textbf{Supplementary Fig.}~\ref{suppfig:7},~\ref{suppfig:8}), while baselines frequently yield over-smoothed outputs as a consequence of severe noise suppression. It is important to note that pixel-wise metrics such as PSNR, SSIM and MS-SSIM tend to disproportionally penalize sharp generative reconstructions for even minor spatial or morphological deviations while favoring blurred predictions. Given the inherent difficulty of this task and the conservative bias of these metrics, the fact that \ourmethod delivers superior visual clarity while still matching the quantitative performance of specialized 3D denoising networks provides strong evidence of its effectiveness in signal recovery and structural preservation. Collectively, these results indicates that, despite being developed as a broadly applicable restoration framework, \ourmethod achieves competitive restoration quality in this demanding regime.

\subsection*{Reliable biological quantification under extreme volumetric sparsity}

The ultimate objective of computational super-resolution microscopy extends beyond visual aesthetics; it fundamentally aims to facilitate robust downstream analysis. In the case of SRS microscopy, cell nuclei segmentation plays a critical role in pathology and intraoperative applications~\cite{Orringer_NatBiomedEng,Hollon_NatMed}, where the shape, size, and distribution of cell nuclei often indicate tissue abnormality. High-resolution SRS imaging is implemented in order to extract those details, at the expense of imaging throughput that is also necessary for imaging larger tissue sections or sampling multiple tissue slices. To evaluate the translational value of \ourmethod, we further analyze the model's generative capacity under the extreme 16x volumetric super-resolution setting (4x downsampling in the lateral plane and 4x along the axial dimension).

\begin{figure}
    \centering
    \includegraphics[width=\textwidth]{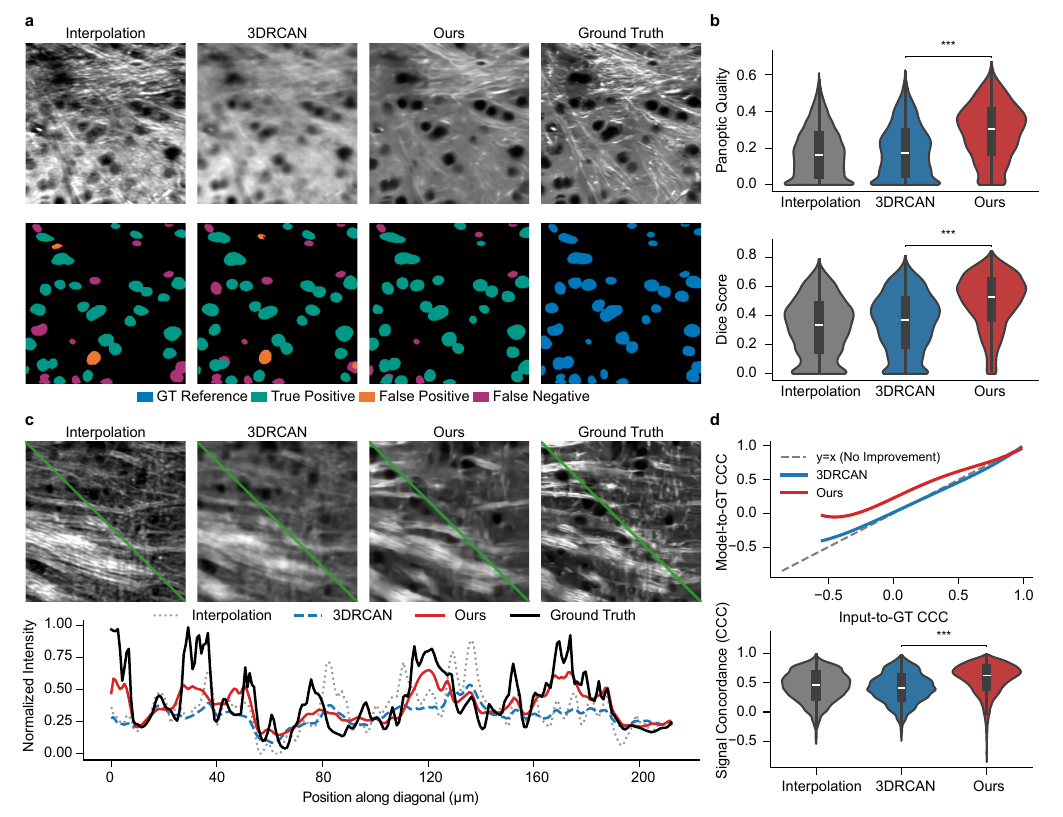}
    \caption{\textbf{Downstream analysis.} \textbf{a, Improvement in segmentation quality.} Comparative evaluation of cell segmentation using Cellpose-SAM. While baseline methods suffer from severe under-segmentation (merging) and false positives, our method yields distinct boundaries, significantly reducing segmentation errors and enabling precise single-cell quantification. 
    \textbf{b, Quantitative improvement of segmentation.} Violin plots show that \ourmethod\ substantially improves segmentation quality relative to the degraded input and all baselines. 
    Despite this challenging evaluation setting, \ourmethod\ shifts the distributions toward markedly better scores and a higher density of well-segmented cases.
    \textbf{c, Restoration of signal intensity profiles.} Diagonal line scans confirm that slice-wise baselines fail to correct signal distortions, merely mimicking the degraded input noise floor. In contrast, our method faithfully restores the signal envelope and recovers peak intensities lost in the input.
    \textbf{d, Quantitative concordance of signal recovery.} Regression curves (upper) of Lin’s Concordance Correlation Coefficient (CCC) for our model versus input demonstrate a consistent shift above reference line ($y=x$). Violin plots (lower) further validate the statistical superiority of our method, showing a significantly tighter distribution of high CCC values compared to baselines. All statistical significance was determined using the Wilcoxon signed-rank test.}
    \label{fig4}
\end{figure}

We first assessed structural integrity via slice-wise segmentation using CellPose-SAM~\cite{cellpose_carsen,cellposesam_pachitariu}. With low-resolution images as input, baseline models (represented by 3DRCAN of the best performance) frequently synthesize over-smoothed slices and overlap artifacts. Such gradient blurring and topological ambiguity introduce both under-segmentation (the erroneous merging of distinct structures) or over-segmentation (fragmented hallucinations) when applying the trained segmentation model. In contrast, \ourmethod successfully mitigates these severe topological distortions (\textbf{Fig.}~\ref{fig4}\textbf{a,b}). By recovering precise anatomical boundaries, our model achieves substantially higher Panoptic Quality Scores~\cite{pqs_kirillov}, Dice scores~\cite{dice_milletari} (\textbf{Fig.}~\ref{fig4}\textbf{b}) and F1 scores (\textbf{Supplementary Fig.}~\ref{suppfig:4}) compared to the baseline. While perfect reconstruction at a 16x sparsity ratio remains an open physical challenge, \ourmethod successfully bridges the gap between severely degraded observations and reliable structural scaffolds, paving the way for automated volumetric segmentation in extreme imaging settings.

Furthermore, we investigated microscopic signal fidelity using 1D line profiles across dense biological structures. As shown in \textbf{Fig.}~\ref{fig4}\textbf{c}, line scans across the diagonal in an axon-dense region reveal that baselines fail to correct signal distortions, exhibiting intensity profiles that merely mimic the degraded input. In contrast, \ourmethod faithfully restores the signal envelope, aligning closely with the Ground Truth distribution and partially recovering peak intensities lost in the low-resolution input. To statistically validate this morphological concordance, we computed the Concordance Correlation Coefficient (CCC)~\cite{ccc_lawrence}, which strictly penalizes both value shifts and scale distortions. The significantly higher CCC scores (\textbf{Fig.}~\ref{fig4}\textbf{d}) as well as Pearson Correlation Coefficient (PCC) and 1D Structural Similarity (1D SSIM) (\textbf{Supplementary Fig.}~\ref{suppfig:5}) achieved by \ourmethod confirm that our architecture robustly recovers lost signals and better preserves the morphological variations required for rigorous biophysical quantification. 

\newpage
\section*{Discussion}

Chemical imaging faces a persistent challenge in balancing resolution, field of view, and imaging speed. As a result, high-resolution volumetric acquisition is often prohibitively slow due to point-scanning and sensitivity-limited imaging physics. One promising strategy to address this challenge is to acquire volumetric data at lower resolution with a larger point spread function and expanded field of view, followed by computational restoration to high resolution, thereby alleviate this tradeoff and substantially increasing the throughput of high-resolution chemical imaging. However, most existing 3D computational models are tailored to axial resolution enhancement or slice interpolation in fluorescence microscopy, rather than 3D reconstruction of high-resolution volumes from low-resolution image acquisitions with enlarged point spread function.  

We developed \ourmethod, a foundation model for 3D super-resolution microscopy and volumetric image restoration, to address this challenge. Driven by a conditional 3D Diffusion Transformer (DiT) with decomposed spatial attention, \ourmethod explicitly accounts for the optical anisotropy inherent in volumetric data while circumventing the prohibitive memory bottlenecks associated with full 3D processing. By internalizing a rich SRS imaging dataset of over 2.5 million spatial slices, our model demonstrates strong capabilities in a range of highly challenging downstream tasks, including extreme 16-fold volumetric super-resolution, ultra-low SNR 3D denoising, and 3D imaging restoration under joint noise and resolution loss (BioTISR). Across these severely degraded regimes—characterized by spatial sparsity or severe photon starvation—\ourmethod overcomes the inherent limitations of conventional restoration methods, faithfully restoring biological signals and anatomical structures with consistent volumetric continuity.

Our approach represents a different modeling paradigm compared to much of the existing computational microscopy literature discussed in the Introduction.  Classical microscopy restoration methods, including denoising, deconvolution, super-resolution, and anisotropy-correction, are typically designed for a specific restoration task within a given imaging regime. In contrast, our work not only addresses a throughput-oriented super-resolution challenge, but also establishes a versatile framework that can be adapted across multiple microscopy modalities and volumetric restoration tasks. Our task also differs fundamentally from recent volumetric interpolation approaches such as InterpolAI. Although Interpolation-based methods can be effective in predicting missing slices, they do not explicitly address signal degradation arising from optical blurring or low-SNR acquisition. Our approach operates on volumetric measurements in which all voxels are physically acquired but degraded by resolution loss and noise, and focuses on restoring high resolution structure by leveraging learned 3D biological priors constrained by the measured data. This distinction leads to significantly better performance of \ourmethod in the most important 3D super-resolution and denoising tasks.

In the context of chemical imaging, particularly Raman and SRS microscope, several computational super-resolution approaches have been explored. However,  existing methods primarily rely on sparsity priors \cite{Lin_LSA}, deconvolution \cite{Shi_NatMethods}, or 2D CNN‑based enhancement \cite{Manifold_OE,Haque_APL}, and are therefore limited to modest resolution improvement, high SNR requirement, and 2D imaging applications. \ourmethod addresses a fundamentally different and more challenging problem: reconstructing high‑resolution 3D volumes from low‑resolution images acquired with a large‑PSF acquisitions under low-SNR conditions. This capability enables substantial gain in imaging throughput without requiring specialized instrumentation, directly supporting applications that demand high-speed, high-resolution, and large-area imaging, such as intraoperative diagnosis and drug screening. Moreover, fine-tuning of the model with a small task-specific dataset is expected to facilitate adaptation of the trained foundation model to diverse cell and tissue contexts and imaging conditions. Finally, \ourmethod can be naturally combined with existing hardware-based super-resolution SRS \cite{Lin_LSA,Ozeki_SciAdv,Aguiar_AdvImaging,Kim_SciAdv} or other advanced chemical imaging modalities, offering a pathway to further enhance spatial resolution using standard microscopes and thereby broadening the accessibility and impact of high-resolution chemical imaging across biomedical and translational settings.   

We note that \ourmethod is not specific to chemical imaging. It is generally applicable to microscopy modalities that are constrained by the tradeoff among resolution, imaging field of view, and speed. Despite its strong performance, \ourmethod presents several important avenues for future improvement. First, the iterative denoising process of diffusion models poses an inference speed bottleneck, which may limit deployment in time-sensitive clinical applications such as rapid intraoperative pathology. Future work could explore integrating advanced sampling acceleration or single-step generative paradigms—such as Mean Flow~\cite{meanflow_geng} or the recently proposed Generative Modeling via Drifting~\cite{drifting_deng}—to achieve real-time 3D decoding in a single forward pass. Second, as with other generative methods, the model carries a residual risk of structural hallucination when applied to rare, out-of-distribution images. An important direction for addressing this issue is the development of 3D uncertainty quantification modules based on diffusion variance to output voxel-wise confidence maps that can delineate safety margins for downstream biological analysis. Finally, while the model exhibits strong generalizability, transitioning to imaging systems with distinct physical principles (e.g., electron microscopy or magnetic resonance imaging) currently requires retraining of the model. Expanding the pretraining corpus to encompass a broader range of imaging modalities may facilitate faster adaptation to restoration tasks with substantially different imaging physics, thus reducing the data-collection burden for new downstream applications.

\newpage 
\section*{Methods}\label{sec:methods}

\subsection*{Details of the pretraining data}\label{method:pretraining}

\textbf{Large-Scale Data Curation.} We curate a massive, depth-spanning dataset comprising approximately 2.38 million high-resolution images ($256 \times 256$ pixels), which consists of roughly 1.81M spectral images and 0.57M spatial images, organized into a mix of single slices, shallow volumes (10-slice cubes), and deep volumes (20-slice cubes). To ensure comprehensive feature coverage, we further include the training splits of three paired benchmarking datasets (detailed in subsequent sections) in the pretraining, contributing an additional 0.17 million slices and bringing the total training corpus to 2.55 million slices. This corpus exposes the model to a representative biological manifold in a scale-invariant and depth-agnostic latent space, establishing a comprehensive structural prior for potential downstream tasks. 

\textbf{Physical Simulation.} Since accurate simulation of axial degradation inherently requires true volumetric context, we selectively apply physical degradation simulations strictly to the 10-slice and 20-slice spatial cubes. For these volumetric subsets, we synthesize corresponding low-quality counterparts by simulating point-spread function (PSF) convolutions and depth-dependent scattering. This targeted simulation strategy ensures that the Diffusion Transformer learns accurate, physically grounded rules of optical degradation.

\textbf{Probabilistic Masked Modeling.} For the remainder of the dataset, we introduce a probabilistic masked modeling paradigm to explicitly elicit structural reasoning. During pretraining, inputs are subjected to a stochastic masking pipeline: with a $90\%$ probability, $50\%$ to $75\%$ of the input patches are dynamically masked; with the remaining $10\%$ probability, an identity mapping (no masking) is applied. The inclusion of identity mapping ensures that the network retains high-fidelity reconstruction capabilities, while the heavy masking acts as a severe information bottleneck. This extreme information bottleneck forces the model to utilize long-range spatial context to deduce missing structures.

The pretraining is then performed within a conditional latent diffusion framework, which is extensively detailed in subsequent sections.

\subsection*{Details of \ourmethod }\label{method:method}
\subsubsection*{Deep Convolutional Feature Encoder} \ourmethod leverages a deep convolutional encoder to extract faithful biological signals. While popular transformer-based architectures excel at capturing global context, convolutional neural networks (CNNs) are intrinsically superior at capturing local, high-frequency signals. By leveraging these strong local inductive biases, this module acts as a structural anchor, explicitly extracting and preserving fine-grained textures and high-frequency boundaries from the degraded inputs to enforce spatial fidelity. 

The primary function of the encoder is to map the low-quality (LQ) input volume into a dense feature representation that perfectly aligns with the diffusion latent space. Specifically, given an LQ input, the convolutional module progressively downsamples the lateral dimensions and encodes the spatial information, yielding a feature sequence with dimensions $D \times H \times W \times C_{latent}$. This strictly matches the dimensions of the VAE-encoded diffusion latent sequence, ensuring a seamless voxel-to-voxel spatial correspondence between the guiding structural conditions and the diffusion latents. 

\subsubsection*{Diffusion Transformer Backbone} In \ourmethod, we adopt a Diffusion Transformer (DiT) backbone for the generative diffusion process. Conventional U-Net architectures often suffer from severe information bottlenecks over long spatial distances due to their reliance on local convolutional routing. 
In contrast, our Transformer-based model possesses inherent global spatial contextualization. By modeling long-range interactions with an $O(1)$ path length, this architecture circumvents the progressive signal dilution typical of deep convolutional hierarchies. Consequently, it excels at routing structural information across distant regions, effectively suppressing localized hallucinations and maintaining biological morphology and anatomical continuity along the depth axis.

Operating directly in the high-dimensional pixel space of 3D volumes is computationally prohibitive. Following the latent diffusion paradigm, we construct the high-resolution targets in a compact latent space of a pretrained Variational Autoencoder (VAE).  Let the compressed 3D latent representation be $z_0 \in \mathbb{R}^{D \times H \times W \times C}$, where $D, H, W$ denote the physical depth, width, and height, respectively, and $C$ denotes the latent channel dimension. 
We formulate the generative process as a conditional denoising objective. Given a timestep $t$ and a comprehensive condition representation $c$ (which encapsulates both global semantics and local spatial anchors), the network $\epsilon_\theta$ is trained to predict the injected noise $\epsilon$ using the standard diffusion objective:
$$
\mathcal{L} = \mathbb{E}_{z_0, \epsilon \sim \mathcal{N}(0, I), t} \left[ \left\| \epsilon - \epsilon_\theta(z_t, t, c) \right\|^2 \right]
$$
The exact mechanisms of how the condition $c$ is extracted and injected into the network are extensively detailed in the following section.

A critical design choice in our diffusion backbone is the tokenization strategy. Unlike naive 3D patchification that flattens the depth dimension, we apply patchification strictly on the lateral dimensions ($H, W$) of the VAE latent space. 

This design is physically and mathematically motivated: standard VAEs inherently encode 2D intra-slice lateral manifolds. Premature 3D tokenization would forcefully entangle these well-structured lateral priors with axial variations before the network's attention mechanisms can explicitly model them. Furthermore, preserving the native depth axis avoids the unnecessary mathematical complexity of axial downsampling and upsampling, providing a structural scaffold for 3D generation.

Specifically, given a patch size of $p$, this operation flattens the lateral axes while natively preserving the depth structure, resulting in a structured latent sequence $S \in \mathbb{R}^{D \times N \times d}$, where $N = (H/p) \times (W/p)$ is the number of tokens per slice and $d$ is the embedding dimension. 
This sequence serves as the direct input to our attention blocks. To efficiently process this $D \times N$ sequence while injecting essential biological inductive biases, we bypass standard full attention and instead employ a decomposed Lateral-Axial attention mechanism, which is elaborated in the following section.

\subsubsection*{Anisotropic Lateral-Axial Attention} Applying standard full-sequence self-attention to our 3D latent sequence $S \in \mathbb{R}^{D \times N \times d}$ imposes a computational complexity of $O((DN)^2)$. This quadratic scaling is computationally intractable for high-resolution volumetric generation. Furthermore, from a biophysical perspective, microscopy imaging is inherently anisotropic. The intra-slice lateral plane (XY) captures high-frequency structural textures, whereas the inter-slice axial dimension (Z) is subject to physical sectioning step sizes and optical scattering, representing low-frequency anatomical continuity. A whole-sequence 3D attention mechanism entangles these distinct physical scales, leading to suboptimal feature representation and cross-dimensional aliasing.

Driven by these physical priors and computational constraints, we design a natively anisotropic $\text{DecomposedAttn}(\cdot)$ module (\textbf{Fig.}~\ref{fig1}c) as a direct drop-in replacement for the vanilla self-attention module. This composite module internally factorizes the representation learning into two orthogonal routing stages. First, it performs intra-slice lateral attention to anchor high-frequency anatomical features within each physical focal plane. Subsequently, it performs inter-slice depth attention along the Z-axis to enforce global perception and axial coherence. The unified forward pass of this operator is formulated as a composite function:
$$\text{DecomposedAttn}(Z) = \mathcal{F}_{axial}\Big( \mathcal{F}_{lateral}(Z) \Big)$$
where $\mathcal{F}_{lateral}$ and $\mathcal{F}_{axial}$ denote the standard multi-head self-attention operations applied independently across the lateral sequence length $N$ and the axial sequence depth $D$, respectively. By orthogonally decomposing the lateral and axial contexts, our unified operator strictly bounds the computational complexity to $O(D N^2 + N D^2)$. 

Within the DiT backbone, this module seamlessly integrates into the standard Transformer block architecture. The output of the decomposed attention is then modulated via the Adaptive Layer Normalization (AdaLN)~\cite{dit_william}:$$Z_{out} = Z_{in} + \text{Modulate}\Big( \text{DecomposedAttn}(LN(Z_{in})), \, c \Big)$$
where $\text{Modulate}$ applies the scale and shift parameters dynamically derived from the condition vector $c$ and $LN$ denotes the layer normalization.

\subsubsection*{Latent Decoder}

After diffusion reconstruction in latent space, the restored latent representation is decoded into pixel space. Although the latent diffusion backbone already produces better volumetric reconstructions than pixel-space restoration baselines, direct decoding with a standard VAE decoder tends to attenuate fine spatial details. To improve spatial fidelity, we augment the decoding stage with a convolution-based refinement module that directly incorporates the low-quality input volume.

Specifically, after latent reconstruction, the latent output is upsampled to image space using pixel-shuffle operations, converting feature maps of spatial size $32 \times 32$ to $256 \times 256$ while preserving the depth dimension. The resulting features are then concatenated channel-wise with the corresponding low-quality input volume. This fused tensor is passed to a refinement decoder adapted from 3DRCAN. Other than the modified input interface, the remaining architecture follows the original 3DRCAN design.

On the 3D SR task, we compared the adapted decoder against both the original VAE decoder and the standard 3DRCAN restoration model. Notably, \ourmethod with the original VAE decoder already substantially outperformed 3DRCAN, indicating that the main performance gain arises from latent diffusion reconstruction rather than the refinement stage alone. The adapted decoder further improved pixel-level fidelity over the original VAE decoder (\textbf{Supplementary Fig.}~\ref{suppfig:6}). Based on these results, we use the adapted decoder as the default decoding module in \ourmethod.

\subsection*{Dual-Conditioning Injection}\label{method:condition}

\subsubsection*{Global-Local Strategy.} A fundamental challenge in applying diffusion models to biological imaging is the risk of severe structural hallucination, wherein networks synthesize morphologically plausible but anatomically fictitious features. To enforce imaging fidelity while maintaining cross-modality adaptability, we design a dual-pathway conditioning system. This system orthogonally decouples global semantic guidance from local structure anchoring.

\subsubsection*{Global Guidance via DINOv2.} To equip the model with a understanding of the input volume across different settings, we extract global semantic priors using a pretrained DINOv2 encoder. We first process the individual slices of the volumetric input independently. Subsequently, we apply global average pooling across all lateral and axial dimensions ($D, H, W$). This operation compresses the feature sequence into a compact semantic embedding. This embedding is subsequently fused with the diffusion timestep embedding $t$ and injected into the diffusion backbone via AdaLN. Operating as a global routing mechanism, this AdaLN modulation dynamically scales and shifts the normalized latent stream, modulating the denoising trajectory.

\subsubsection*{Local Structure Anchoring.} While DINOv2 embeddings provide global guidance, it lacks the high-frequency spatial precision required for dense biological structure reconstruction. To establish a rigid structural scaffold, we explicitly anchor the generative process using the features extracted by the Deep Convolutional Encoder. These local features are integrated with the noisy latent $z_t$ via channel-wise concatenation at the input layer, which provides the diffusion model with a spatial scaffold of the biological structures from the very onset of the denoising process.

Furthermore, a recognized vulnerability in deep generative networks is condition dilution. Initial spatial constraints concatenated at the input layer are susceptible to signal decay across successive blocks.  To prevent the attenuation of high-frequency signals, we introduce a block-wise condition injection. Specifically, within each DiT block, the convolutional local condition $c_{local}$ is persistently re-injected into the latent stream $z$:

$$z = z + \text{LN}(\sigma(\text{Linear}(c_{local})))$$

where $\text{LN}$ denotes Layer Normalization and $\sigma$ represents the non-linear activation function. Because the tensor topology of $c_{local}$ is strictly aligned with the latent stream $x_{in}$ across all lateral and axial dimensions, this direct element-wise addition exploits their explicit spatial correspondence.

\subsection*{Competing methods}
We compared \ourmethod\ against five representative restoration baselines: Super-Resolution Convolutional Neural Network (SRCNN)~\cite{srcnn_dong}, Content-Aware Image Restoration (CARE)~\cite{care_weigert}, Image Restoration Using Swin Transformer (SwinIR)~\cite{swinir_liang}, Residual Local Attention Network (RLN)~\cite{rln_li}, and 3D Residual Channel Attention Network (3DRCAN)~\cite{3drcan_chen}. SRCNN was included as a classical convolutional super-resolution model that provides a simple and widely recognized reference for image restoration. CARE was selected as a strong microscopy-oriented baseline that has been extensively used for denoising and restoration in biological imaging. SwinIR was included as a modern transformer-based restoration model with strong performance across a broad range of image restoration tasks. RLN and 3DRCAN were further included as strong restoration baselines with competitive performance in volumetric image restoration. Together, these methods span classical convolutional restoration, microscopy-specific supervised restoration, transformer-based image reconstruction, and dedicated volumetric restoration architectures. For volumetric tasks, 2D baselines were applied in a slice-wise manner, which reflects their native formulation. Although some 2D baselines can be adapted to pseudo-3D formulations by stacking adjacent slices as input, we did not include these variants because such adaptations rely on auxiliary input assumptions that are not naturally matched to our benchmarks and do not provide a directly comparable formulation.

We did not include several other existing 3D microscopy models as direct baselines because their primary task settings differ substantially from ours. In particular, many volumetric methods are developed for axial correction or anisotropy restoration, where the goal is to improve low-quality axial views toward lateral image quality, often in unsupervised or weakly supervised settings. 
Other methods focus on interpolation between sparsely sampled planes to recover unobserved slices. By contrast, our primary setting is supervised, throughput-oriented volumetric restoration from physically acquired but severely degraded measurements. Because these methods address different reconstruction objectives and rely on substantially different input assumptions, we did not regard them as directly comparable baselines for the present benchmarks.

\subsection*{Details of downstream evaluation and benchmarks}

\textbf{SRS Data: Tissue Samples.} Whole mouse brain coronal sections (sourced from the University of Washington Animal Use Training Services) were acquired using 10$\times$ and 30$\times$ objectives. For the 10$\times$ acquisitions, an axial step size of 2 microns was employed (these oversampled volumes were downsampled to achieve a 4-fold axial sparsity later). For the 30$\times$ acquisitions, a step size of 1 micron was used. After excision, bulk tissues were chemically fixed by 4\% paraformaldehyde and stored in phosphate-buffered saline (PBS). Thin sections (300 microns) of tissue were prepared by vibratome sectioning and then placed on a glass slide with additional PBS. To prepare the sample for imaging, a glass coverslip was placed on top to sandwich the tissue specimen between the glass slide and the glass coverslip using double-sided tape as a spacer. 

\textbf{SRS Data: Cell Culture.} The human lung adenocarcinoma A549 cell line (ATCC) was maintained at 37 °C in an atmosphere of 5\% CO2 (v/v) humidified incubator. A549 cells were cultured in DMEM (Gibco, 11965092) supplemented with 10\% fetal bovine serum (Hyclone) and 1\% penicillin–streptomycin (Fisher Scientific).

\textbf{SRS Data: Acquisition.} To rigorously evaluate the performance of 3D volumetric generation, we constructed two perfectly aligned, large-scale benchmarking datasets for super-resolution and denoising. All raw volumetric data were acquired using SRS microscopy. 
Briefly, the system (FLINT FL2, Light Conversion) outputs a 1030 nm beam with a 77 MHz (f0) repetition rate. This beam is split by a polarizing beam splitter into two arms, one arm is sent to an optical parametric oscillator (OPO) to generate the wavelength tunable pump beam (790 nm), and the other acts as a Stokes beam (1030 nm). The Stokes beam is modulated at 19.25 MHz (f0/4) by an electro-optical modulator (EOM). Both the Stokes and pump beams are chirped individually by a grating-based pulse stretcher to achieve a spectral resolution of 16 cm-1 for DMSO solvent. The two beams are combined with a dichroic mirror and then sent to an inverted laser scanning microscope. Images were taken with either a 10x, 0.45 NA objective (Nikon, CFI Plan Lambda D 10x) with 0.52 NA condenser (Nikon, LWD) or a 30x, 1.05 NA objective (Olympus, UPlanSApo) with 1.4 NA condenser (Nikon, Achr-Apl). All images are collected at room temperature and were acquired at the lipid (2850 cm-1) and protein (2930 cm-1) peaks separately.
For image pairs at high- and low-signal, powers are set to 40 mW pump (790 nm) and 60 mW Stokes (1030 nm) at the focus for imaging at high signal and powers were set to 40 mW pump (790 nm) and 15 mW Stokes (1030 nm) for low signal. Powers were automatically changed by using a servo motor to alter the halfwave plate rotation on the Stokes arm. 
For image pairs that involve XYZ image stacks, sample focus was determined by highest intensity after scanning the spatial z by a piezo objective stage (nPFocus400). A custom MATLAB script was used to perform XYZ mosaic scans at different powers for dataset generation. 
To maintain the biological relevance of the evaluation and exclude regions dominated by scattering and signal depletion, all compiled volumes were strictly constrained to a maximum imaging depth of 60 microns.

\textbf{SRS Data: Image registration.} 2D image pairs are taken in sequence and did not require registration. 3D image pairs for high- and low-resolution datasets were taken as large separate mosaic scans with the low-resolution 10x scans occurring first. Images were taken such that one 10x tile corresponds to nine tiles of the high-resolution, 30x condition. On average, 70 tissues were imaged by the 10x objective for 9 hours; imaging the corresponding areas with the 30x objective took 62 hours. Images were then registered through SIFT for feature extraction and RANSAC for outlier removal. After image matching, each LR and HR pair is cropped to a 256x256 patch in xy for model training.

\textbf{SRS Data: Splitting.} To strictly prevent potential data leakage, our dataset splitting was executed exclusively at the Region of Interest (ROI) level with a ratio of 7:1:2. Because imaging in different ROIs naturally varies in valid depth, the resulting volumetric cube counts exhibit slight numerical deviation of the above ratio. This spatial segregation ensures that the testing sets consist of entirely unseen physical micro-environments, providing a completely unbiased evaluation of reconstruction. \ourmethod was finetuned on the training split and evaluated on the held-out test split.

\textbf{SRS Data: Processing.}
For the SRS benchmarks, raw volumetric TIFF data were converted into paired samples for both 3D super-resolution and 3D denoising. All images were laterally normalized using 1st/99th percentile clipping within each slice and rescaling to $[0,1]$. High-Quality (HQ) targets were further denoised with carefully parameterized BM3D to mitigate residual shot noise inherent to high-power acquisition.

For 3D super-resolution, matched high-resolution and low-resolution acquisitions were used to construct paired volumes with a 4$\times$ difference in axial sampling density. HQ targets retained the original dense z-stack, whereas LQ inputs were generated by axial downsampling (raw LQ data were originally oversampled). Training cubes were extracted with HQ depth 20 and LQ depth 5; for slice-wise benchmarking, LR stacks were trilinearly interpolated along z-axis to match the HQ depth.

For 3D denoising, the low-power acquisition was used as the LQ input and the high-power acquisition as the HQ target. Paired 3D cubes of depth 20 were extracted with overlapping sliding windows (stride 10). Corresponding 2D slice pairs were also stored for slice-wise evaluation. 

\textbf{SRS 3D Super-Resolution Benchmark.} For the volumetric super-resolution task, the spatially segregated ROIs were cropped into standardized 3D cubes with dimensions of 20 × 256 × 256 (Depth × Height × Width). The final benchmark comprises approximately 4,490 cubes for training, 640 for validation, and 1,285 for independent testing.

\textbf{SRS 3D Denoising Benchmark.} For the volumetric denoising task, we compiled an independent dataset strictly matched in dimensions (20 × 256 × 256 cubes). To systematically evaluate algorithmic robustness against depth-dependent optical degradation, this benchmark incorporates two different signal-to-noise ratio (SNR) settings. The dataset (partitioned into 2,352 training, 336 validation, and 672 testing cubes) is evenly distributed between two distinct physical degradation regimes: a medium-signal condition (SNR $\le$ 40) typically found in shallow tissues, and a low-signal condition (SNR $\le$ 2) representing photon starvation in deep tissues or extreme external disturbance.

\textbf{Evaluation on BioTISR}
BioTISR dataset contains 4D (including temporal dimension) SIM data from three biological structures: outer mitochondrial membrane, microtubules, and F-actin. For each sample type, the dataset provides independent regions of interest (ROIs), acquired using the 3D-SIM mode of a Multi-SIM system under different illumination levels. Each ROI contains paired low- and high-resolution volumetric observations stored in MRC format, including a reconstructed high-resolution SIM volume and a corresponding lower-resolution widefield input.

\textbf{BioTISR Data: Splitting.} We used the 3D subset only and performed dataset splitting at the ROI level to avoid information leakage across train, validation, and test sets. Within each biological category, ROIs were randomly divided into 70\% training, 10\% validation, and 20\% testing subsets. \ourmethod was finetuned on the training split and evaluated on the held-out test split.

\textbf{BioTISR Data: Processing}
For preprocessing, each raw volume of shape (H,W,10×k) was first rearranged into (10,k,H,W), separating the temporal dimension (10 frames) from the axial dimension (k slices, which varies). We construct the cube along the axial dimension for each fixed time point as it is  consistent with the volumetric restoration setting studied in other experiments. We standardized the cube depth to 8 slices and bicubically upsample lateral planes in low-quality volume from 512 × 512 to 1024 × 1024 to match the spatial size of the corresponding high-resolution targets. To reduce edge-related artifacts and standardize the evaluation field, we extracted four 256 × 256 patches from the central 512 × 512 region of each volume using a stride of 256, yielding input and target cubes of shape (8,256,256). Finally, each 2D slice was independently normalized using percentile normalization. Specifically, intensities were clipped to the [1,99] percentile range and linearly rescaled to [0,1]. High-Quality (HQ) targets were further denoised with carefully parameterized BM3D to mitigate noise.



\printbibliography

\newpage

\appendix
\renewcommand{\figurename}{Supplementary Figure}
\setcounter{figure}{0}
\renewcommand{\tablename}{Supplementary Table}

%

\begin{figure}
    \centering
    \includegraphics[trim=15mm 0 15mm 0,, width=0.7\textwidth]{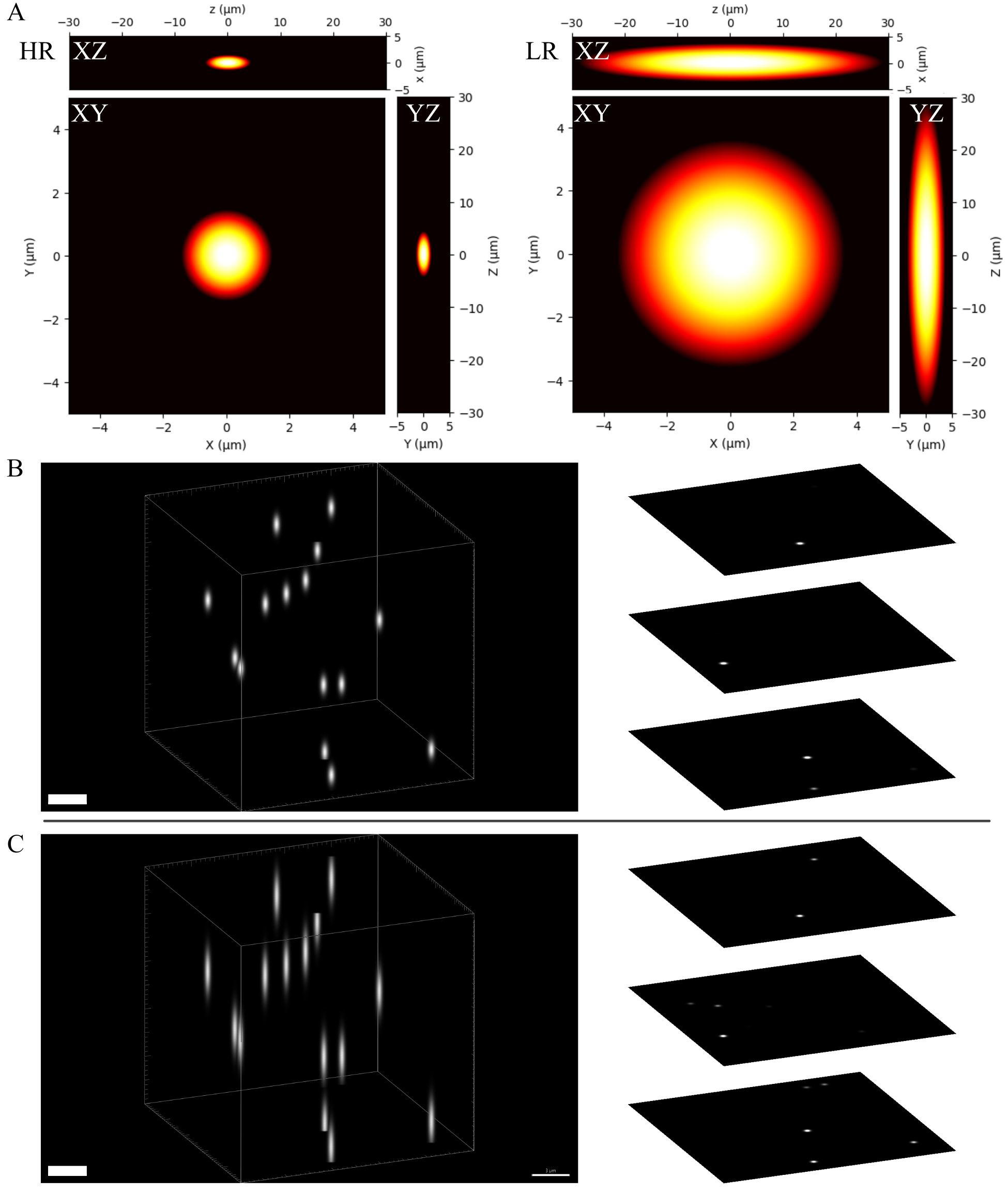}
    \caption{ \textbf{Point spread function illustration.} \textbf{A, Differences in the point spread function between high-resolution (HR) and low-resolution (LR) objectives.} The LR objective has a more spatially dispersed kernel, yielding a broader effective collection profile and greater axial signal entanglement. This causes each voxel to contain more mixed volumetric information and makes ideal slice-wise reference or alignment between LR and HR acquisitions difficult. \textbf{B, C, Simulated examples of this effect.} \textbf{B,} simulated 3D representation of point sources acquired with the HR objective, with cross-sections from the top, middle, and bottom shown on the right. \textbf{C,} simulated 3D representation of the same objects using LR acquisition parameters. Scale bar 5 microns.}
    \label{suppfig:1}
\end{figure}


\clearpage
\begin{figure}
    \centering
    \includegraphics[trim=15mm 0 15mm 0,, width=0.5\textwidth]{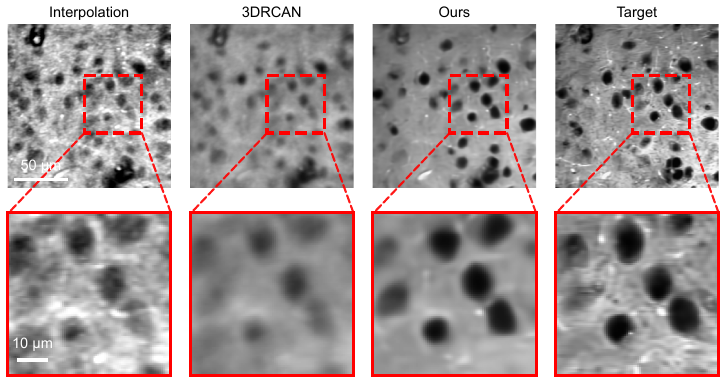}
    \caption{\textbf{Visual comparison of super-resolution quality.} Representative images of the low-quality input, 3DRCAN, our method, and the high-quality target, with enlarged views shown below. Our reconstruction shows clearer nuclear appearance and better-defined biological structure boundaries.}
    \label{suppfig:2}
\end{figure}

\clearpage

\begin{figure}
    \centering
    \includegraphics[trim=15mm 0 15mm 0,, width=0.5\textwidth]{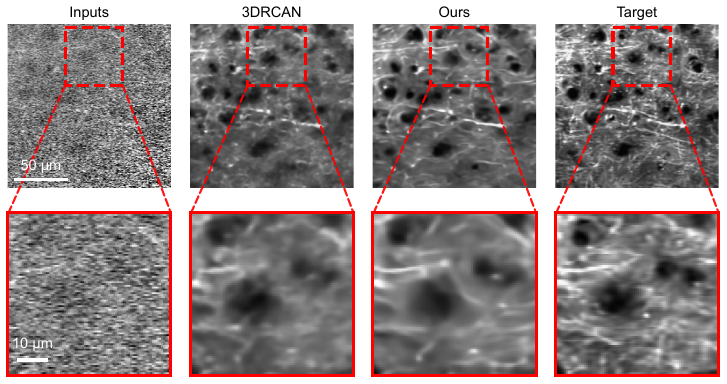}
    \caption{\textbf{Visual comparison of denoising quality.} Representative images of the low-quality input, 3DRCAN, our method, and the high-quality target, with enlarged views shown below. Our reconstruction shows improved structural clarity and better-defined biological boundaries.}
    \label{suppfig:3}
\end{figure}

\clearpage

\begin{figure}
    \centering
    \includegraphics[trim=15mm 0 15mm 0,, width=0.5\textwidth]{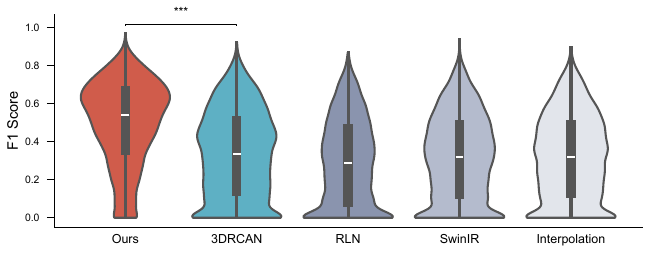}
    \caption{ \textbf{Cell segmentation performance measured by F1 score.} Quantitative comparison of segmentation quality across different reconstruction methods. Our method achieves the highest F1 score among all methods. Statistical significance was determined using the Wilcoxon signed-rank test.}
    \label{suppfig:4}
\end{figure}

\clearpage

\begin{figure}
    \centering
    \includegraphics[trim=15mm 0 15mm 0,, width=0.5\textwidth]{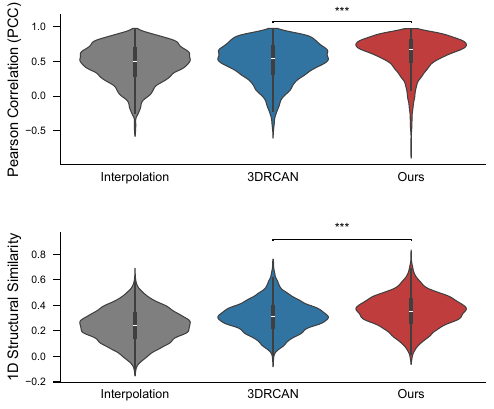}
    \caption{\textbf{PCC and 1D SSIM of reconstructed signal profiles.} \textbf{a,} PCC between reconstructed and target intensity profiles. \textbf{b,} 1D SSIM between reconstructed and target intensity profiles. Our method shows the highest profile consistency with the target. Statistical significance was determined using the Wilcoxon signed-rank test.}
    \label{suppfig:5}
\end{figure}

\clearpage

\begin{figure}
    \centering
    \includegraphics[trim=15mm 0 15mm 0,, width=\textwidth]{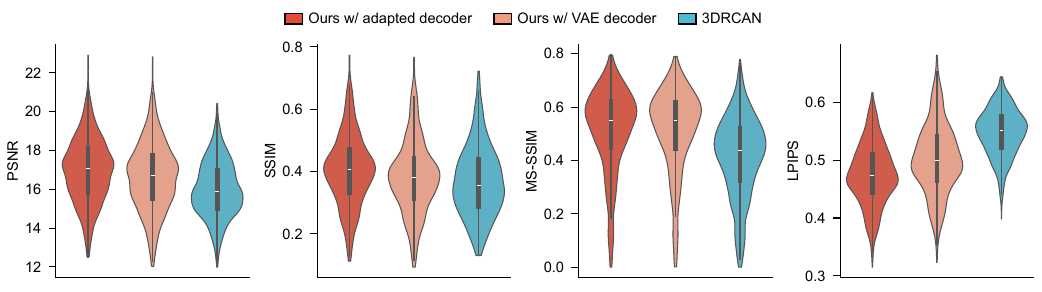}
    \caption{\textbf{Performance comparison of different decoders and 3DRCAN.} Comparison of PSNR, SSIM, MS-SSIM, and LPIPS for \ourmethod\ with the adapted decoder, \ourmethod\ with the original VAE decoder, and the standard 3DRCAN restoration model. Notably, \ourmethod\ with the original VAE decoder already substantially outperforms 3DRCAN. The adapted decoder further improves reconstruction quality over the original VAE decoder.}
    \label{suppfig:6}
\end{figure}

\clearpage
\begin{figure}
    \centering
    \includegraphics[trim=15mm 0 15mm 0,, width=0.5\textwidth]{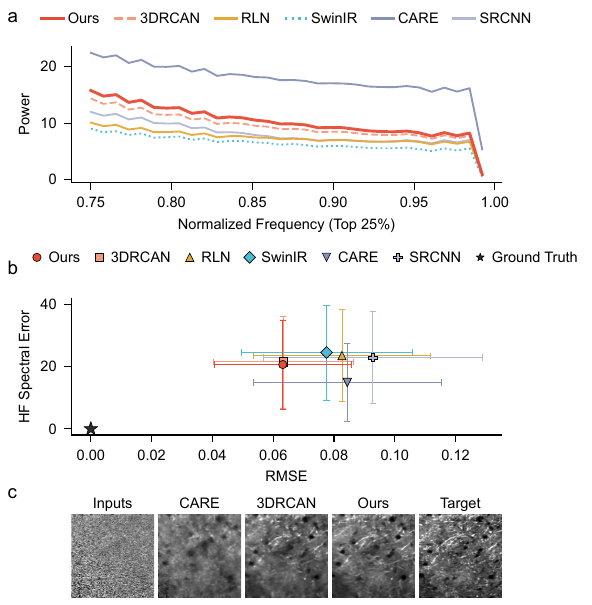}
    \caption{\textbf{Spectral analysis and perception-distortion trade-off.} \textbf{a, High-frequency radial power spectrum.} Averaged power spectrum of the top 25\% frequencies (normalized by Nyquist frequency). Our method resolves sharper anatomical features by retaining substantially higher spectral energy than most baselines. The ground truth is omitted to preserve the visualization scale due to its significantly higher energy. The higher energy observed in CARE reflects noise artifact, evidenced by its high voxel-wise error. \textbf{b, Perception-distortion evaluation.} Scatter plot of High-Frequency (HF) Spectral Error versus Root Mean Square Error (RMSE). HF Spectral Error is defined as the average of absolute power difference with the ground truth. Our approach achieves the best voxel-wise fidelity while maintaining the most competitive HF spectral error, positioning it closest to the ideal Ground Truth (black star). Error bars indicate standard deviation across the test volumes. \textbf{c, Visual comparison between CARE, 3DRCAN and \ourmethod.} CARE fails to fully remove gaussian noise, due to its residual connection and limited denoising capability in extreme scenario, which explain the abnormal high-frequency power in its generation. \ourmethod provides better fine-grained details in comparison with 3DRCAN.}
    \label{suppfig:7}
\end{figure}

\begin{figure}
    \centering
    \includegraphics[trim=15mm 0 15mm 0,, width=0.2\textwidth]{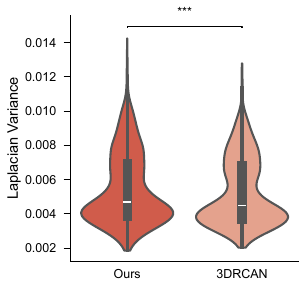}
    \caption{\textbf{Comparison of Laplacian variance.} Our method resolves sharper anatomical features by retaining substantially higher Laplacian variance than the representative baseline. Statistical significance was determined using the Wilcoxon signed-rank test.}
    \label{suppfig:8}
\end{figure}
 



\end{document}